%% file: main.tex
\documentclass[letterpaper]{article} 
\usepackage{aaai2026}  
\usepackage{times}  
\usepackage{helvet}  
\usepackage{courier}  
\usepackage[hyphens]{url}  
\usepackage{graphicx} 
\usepackage{natbib}  
\usepackage{caption} 
\usepackage{algorithm}
\usepackage{algpseudocode}
\usepackage{amsmath} 
\usepackage{amssymb}
\usepackage{amsthm}         
\usepackage{framed}   
\usepackage{breakcites}
\usepackage{subcaption}  
\usepackage{comment}
\usepackage{makecell}
\usepackage{placeins}
\usepackage{booktabs}
\usepackage{multirow}
\usepackage{siunitx}

\newcommand{\norm}[1]{\left\lVert#1\right\rVert} 
\newtheorem{definition}{Definition}
\newtheorem{theorem}{Theorem}
\newtheorem{proposition}{Proposition}
\newtheorem{corollary}{Corollary}
\usepackage{newfloat}
\usepackage{listings}
\DeclareCaptionStyle{ruled}{labelfont=normalfont,labelsep=colon,strut=off} 
\floatstyle{ruled}
\newfloat{listing}{tb}{lst}{}
\floatname{listing}{Listing}
\title{Discrete Optimization of Min–Max Violation and its Applications Across Computational Sciences}
\author{
    Cheikh Ahmed\textsuperscript{\rm 1},
    Mahdi Mostajabdaveh\textsuperscript{\rm 1}\thanks{Corresponding Author : mahdi.mostajabdaveh1@huawei.com },
    Samin Aref\textsuperscript{\rm 2},
   Zirui Zhou\textsuperscript{\rm 1}
}
\affiliations{
    \textsuperscript{\rm 1}Huawei Technologies Canada, Burnaby, Canada\\
    \textsuperscript{\rm 2}Department of Mechanical and Industrial Engineering
University of Toronto, Toronto, Canada\\

}

\begin{document}

\maketitle


\begin{abstract}

We introduce the Discrete Min–Max Violation (DMMV) as a general optimization problem which seeks an assignment of  discrete values to variables that minimizes the largest constraint violation. This context-free mathematical formulation is applicable to a wide range of use cases that have worst‐case performance requirements.
After defining the DMMV problem mathematically, we explore its properties to establish a foundational understanding. To tackle DMMV instance sizes of practical relevance, we develop a GPU-accelerated heuristic that takes advantage of the mathematical properties of DMMV for speeding up the solution process. We demonstrate the versatile applicability of our heuristic by solving three optimization problems as use cases: (1) post-training quantization of language models, (2) discrete tomography, and (3) Finite Impulse Response (FIR) filter design.
In quantization without outlier separation, our heuristic achieves 14\% improvement on average over existing methods. In discrete tomography, it reduces reconstruction error by 16\% under uniform noise and accelerates computations by a factor of $6$ on GPU. For FIR filter design, it nearly achieves 50\%  ripple reduction compared to using the commercial integer optimization solver, Gurobi. Our comparative results point to the benefits of studying DMMV as a context-free optimization problem and the advantages that our proposed heuristic offers on three distinct problems. Our GPU-accelerated heuristic will be made open-source to further stimulate research on DMMV and its other applications. The code is available at \url{https://anonymous.4open.science/r/AMVM-5F3E/}
\end{abstract}

\section{Introduction}

Several real‐world engineering and Artificial Intelligence (AI) applications demand minimization of the worst‐case error over a finite set of choices. For example, post‐training quantization of Large Language Models (LLMs) involves choosing between discrete quantized weight and/or activation values \cite{zhu2024survey,
lang2024comprehensive,jin2024comprehensive}; discrete tomography reconstructions are based on fixed grey‐level intensities \cite{Dart}; and Finite Impulse Response (FIR) filter design enforces a limited set of coefficient values when the hardware is constrained \cite{oppenheim_schafer, Rabiner}.

Classically, worst‐case performance is addressed by continuous minimax $\ell_\infty$ approximation \citep{Continuous_minimax1, Continuous_minimax2}. Given a system of linear equations $Ax \approx b$, the Chebyshev criterion seeks
$
  \min_{x \in \mathbb{R}^n} \|Ax - b\|_\infty,
$
i.e., minimizing the magnitude of the largest residual error (maximum violation). This framework underpins a wide range of computational problems that can be explored from a continuous optimization standpoint. They include designing digital filters \cite{remez}, developing robust statistical models \cite{Barrodale, yi_Linf}, as well as other applications in mathematics \cite{GILLIS2019367, AAM}, image processing \cite{Kahl2005}, and signal processing \cite{PARR}.

When decision variables are to be chosen from a discrete set (e.g., integer values), continuous optimization theory no longer applies. 
Despite the ubiquity of discrete-valued decisions, the underlying optimization problem has received only ad-hoc treatments. Existing works either (i) relies on continuousness and deals with discreteness in a post-processing step\cite{Fir}, or (ii) uses handcrafted task-specific heuristics\cite{Jensen2001}.   
To the best of our knowledge, the literature lacks both a systematic, problem-agnostic analysis of discrete minimax approximation and a scalable, general-purpose algorithm to solve this problem. We therefore introduce the \textbf{Discrete Min‐Max Violation} (DMMV) problem as the natural extension of Chebyshev approximation to discrete domains and propose a competitive solution method for it.

\begin{figure*}[htp!]
  \centering
  \includegraphics[width=0.75\textwidth]{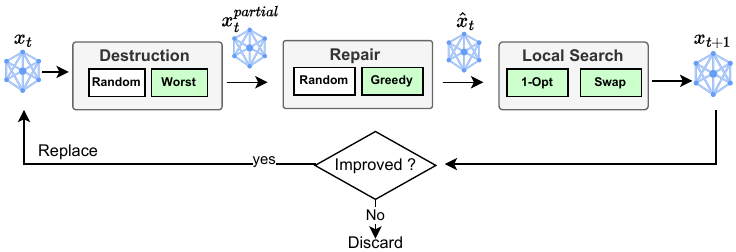}
  \caption{One iteration in our AMVM heuristic framework for solving a large-scale DMMV instance. The green components use GPU parallelization.}
  \label{fig:AMVM}
\end{figure*}

\paragraph{Contributions} This paper makes three key contributions. First, we provide a rigorous and context-free definition of the DMMV problem and establish its NP‐hardness. Second, we propose a GPU-accelerated heuristic, \textbf{Accelerated Maximum Violation Minimizer (AMVM)},  based on the adaptive large neighborhood search \cite{ALNS} tailored for scalable solution of DMMV instances (Figure~\ref{fig:AMVM}). Third, we empirically validate the relevance of our approach using comparison with use-case-specific baseline methods from three domains of application: (1) post‐training quantization of LLMs for edge devices \cite{zhu2024survey}, discrete tomography \cite{Dart}, and FIR filter design \cite{oppenheim_schafer}. 
\paragraph{Main Results} Our AMVM heuristic achieves 14\% improvement on average over existing methods for quantization without outlier separation. In discrete tomography, it reduces reconstruction error by 16\% under uniform noise and accelerates computations by a factor of $6$ on GPU. For FIR filter design, it nearly achieves 50\%  ripple reduction compared to the commercial discrete optimization solver, Gurobi \cite{gurobi}. Background and a brief literature review for each of these three use cases are provided after describing our solution approach.

\section{Mathematical Problem Statement}\label{sec:problem}

The discrete Min-Max violation problem seeks to find values for vector $\mathbf{x}$ out of the discrete set $\mathbb{V}$ to minimize the maximum absolute difference between $\mathbf{Ax}$ and a target vector $\mathbf{b}$. This is formally expressed using the $\ell_{\infty}$-norm  as:
$$ \min_{\mathbf{x} \in \mathbb{V}^n} \norm{\mathbf{Ax} - \mathbf{b}}_{\infty} $$
where $A \in \mathbb{R}^{m \times n}$ is a real-valued matrix, $\mathbf{b} \in \mathbb{R}^m$ is a real-valued vector, and $\mathbf{x} \in \mathbb{V}^n$ is the vector of decision variables.

While the $\ell_{\infty}$-norm objective is non-linear, the DMMV problem can be reformulated as a \textbf{Mixed-Integer Linear Program (MILP)}. This reformulation is obtained by introducing a single, non-negative continuous variable, $t \in \mathbb{R}$, which represents the value of the maximum violation. The problem then reduces to:
\begin{align*}
\min_{\mathbf{x}, t} \quad & t \\
\text{s.t.} \quad & \mathbf{Ax} - \mathbf{b} \le t \mathbf{1} \\
                 & \mathbf{Ax} - \mathbf{b} \ge -t \mathbf{1} \\
                 & \mathbf{x} \in \mathbb{V}^n, \quad t \ge 0.
\end{align*}

\begin{theorem}
The Discrete Min-Max Violation (DMMV) problem is NP-hard.
\end{theorem}

To prove this theorem, we analyze the complexity of the corresponding decision problem, defined as follows.

\begin{definition}[The Decision Version of DMMV]
Given a matrix $A \in \mathbb{R}^{m \times n}$, a vector $b \in \mathbb{R}^m$, a finite set of allowed values $\mathbb{V} \subset \mathbb{R}$, and a threshold $t \in \mathbb{R}$, the decision problem $\text{DMMV}_{\infty}$ asks if there exists a vector $x \in \mathbb{V}^n$ such that $\|Ax - b\|_{\infty} \le t$.
\[
\textsc{DMMV}_\infty = \bigl\{(A,b,\mathbb{V},t)\;\bigm|\;\exists\,x\in \mathbb{V}^n:\|Ax - b\|_\infty \le t\bigr\}.
\]
\end{definition}

We now prove that $\text{DMMV}_{\infty}$ is NP-complete. The NP-hardness of the original DMMV optimization problem follows directly from this result.

\begin{proof}
\label{proof:np_hard}

\textbf{(1) Membership in NP.}
Given an instance \((A\in\mathbb{R}^{m\times n},\,b\in\mathbb{R}^m,\,\mathbb{V}\subset\mathbb{R},\,t\in\mathbb{R})\) and a certificate \(x\in \mathbb{V}^n\), we compute
$r = Ax - b
\quad\text{in }O(mn)\text{ time}$.
Then, we compute \(\max_{1\le i\le m}|r_i|\) in \(O(m)\), and check \(\max_i|r_i|\le t\).  Thus, \(\textsc{DMMV}_\infty\in\mathrm{NP}\).

\textbf{(2) NP‐hardness by reduction from Subset‐Sum.}
Recall the Subset‐Sum problem\cite{subsetsum}: \\
Given $a_1,\dots,a_n\in\mathbb{Z}_{>0},\,S\in\mathbb{Z}_{>0},$  decide if $\sum_{i\in I}a_i = S\text{ for some }I\subseteq\{1,\dots,n\}.$
Construct an instance of \(\textsc{DMMV}_\infty\) as follows:
$
m = 1,\quad D = \{0,1\},\quad A = [a_1\;\;a_2\;\dots\;a_n]\in\mathbb{R}^{1\times n},
\quad b = [S]\in\mathbb{R},\quad t = 0.
$
Then for any \(x\in\{0,1\}^n\),
\[
\|Ax - b\|_\infty \le 0
\;\Longleftrightarrow\;
Ax = b
\;\Longleftrightarrow\;
\sum_{i=1}^n a_i x_i = S.
\]
Hence Subset‐Sum has a solution if and only if \(\|Ax-b\|_\infty\le0\).  This reduction runs in \(O(n)\) time, proving NP‐hardness.

\textbf{(3) Conclusion.}
We showed that the decision version of $\textsc{DMMV}_\infty$ (i.e., asking whether there exists a solution with objective value $\le k$) is in NP and NP‐hard, and hence NP‐complete. It follows that the corresponding DMMV optimization problem is NP‐hard.
\end{proof}

\section{Solution Approach}
\label{sec:solution}

To address large-scale instances of the DMMV problem, we propose AMVM, a GPU-accelerated heuristic that exploits DMMV's underlying mathematical structure to restrict the optimization search space.  

Our proposed improvement heuristic requires an initial solution. This can be obtained by solving the linear-programming relaxation of the DMMV formulation and rounding each variable $x$ to its nearest value from the discrete set $\mathbb{V}$. We then iteratively refine this solution through three phases. 1) \textbf{Destruction}: remove a subset of assignments from the current solution. 2) \textbf{Repair}: restore feasibility by reassigning the removed elements. 3) \textbf{Local search}: apply neighborhood moves to further improve the objective. The algorithm iterates over these three phases until a predetermined stopping criterion is met. We use two destroy operators and two repair operators. At the beginning of each iteration, one destroy-repair pair gets selected adaptively. We also employ two local-search moves, which are applied sequentially to the solution in each iteration. Figure~\ref{fig:AMVM} provides an overview of our heuristic framework.

Additional technical details of our proposed heuristic are provided in the Materials and Methods Section. In the following three Sections, we describe the three use cases to which the DMMV problem and AMVM heuristic are directly applicable. For each use case, comparative results are provided to demonstrate the practical relevance of our method.

\section{LLM Quantization W/O Outlier Separation} 
\label{sec:LLM}

\paragraph{Model compression and quantization}
Large language models tend to use numerous high-precision parameters \cite{BERT}, which leads to high computational costs during inference. A widely used approach to reduce memory and energy usage during inference is to compress a model by replacing high-precision parameters with quantized values of lower precision \cite{zhu2024survey}. This type of model compression is referred to as model quantization \cite{lang2024comprehensive}. The goal is to obtain a resource-efficient model with quantized weights and/or activations whose performance is comparable to the original model \cite{lang2024comprehensive}. Besides addressing the environmental concerns surrounding the intensive resource usage of LLMs \cite{cao2025saliency}, quantization makes model storage and deployment easier and cheaper, especially in resource-constrained environments like edge devices. Most existing quantization methods quantize the parameters with the exception of a subset of them which are called outliers \cite{jin2024comprehensive}. The outliers, more generally known as salient weights \cite{cao2025saliency}, are typically separated and left unquantized as an attempt to restrict the quantization range and limit the performance degradation of the model. The outliers are sometimes a considerable fraction of the parameters. For example, the method known as PB-LLM \cite{shang2023pb} achieve competitive performance, by storing up to 30\% of all weights as salient weights and producing a mixed-precision model.

\paragraph{Drawbacks of quantization with outlier separation}
While quantization methods that involve outlier separation may achieve near-lossless compression down to 4 and 3 bits \cite{kim2023squeezellm,wang2024fp4}, keeping outliers in high precision has practical drawbacks. These
include: (1) memory bandwidth overhead, (2) complexity in memory handling, and (3) hardware requirements to support the computation. (1) Having high-precision outliers effectively increases the average bit-width and undermines the bandwidth savings that quantization is supposed to provide. (2) It produces a mixed-precision model that requires costly unstructured mixed-precision operations during inference \cite{shang2023pb,huang2024billm}. Mixed-precision tensors need custom layouts and packing/unpacking logic, making memory access and kernel design substantially more complex \cite{koo2024opal}. (3) Outliers typically need higher-precision arithmetic (e.g., floating point). This increases design and runtime complexity \cite[Table 11]{xiao2023smoothquant}. Note that the hardware that is suitable for quantization with outlier separation must support both integer and floating-point (FP) computation. Some edge devices, such as Google's edge TPU announced in 2018, do not support FP16 or FP32 arithmetic. Paradoxical to the concept of model quantization that is often motivated by AI inference on edge devices \cite{jin2024comprehensive}, mixed-precision quantized model (that have FP16 outliers) are impossible to run on some edge processors that are designed for AI acceleration.

\paragraph{Quantization without outlier separation as DMMV}
As a use case for demonstrating the applicability of our proposed method and motivated by the earlier arguments, we focus on the post-training quantization of weights without outlier separation. 

Let $W_l\in\mathbb R^{n\times d}$ be the full‑precision weight matrix of layer $l$, whose $r$‑th row is denoted by $w_{l,r}\in\mathbb R^d$. Let $X_l\in\mathbb R^{m\times d}$ be a calibration matrix.  Over a fixed quantization alphabet $Q\subset\mathbb R$, we seek a quantized weight matrix $\widehat W_l\in Q^{n\times d}$ that minimizes the worst‑case entry-wise error between the quantized and original outputs. Concretely,

\begin{align*}
\widehat W_l 
&= \arg\min_{W\in Q^{n\times d}}
   \bigl\lVert W\,X_l^T - W_l\,X_l^T \bigr\rVert_\infty \\[6pt]
&= \arg\min_{W\in Q^{n\times d}}
   \max_{1 \le r \le n}\;
   \bigl\lVert w_r\,X_l^T - w_{l,r}\,X_l^T \bigr\rVert_\infty.
\end{align*}

Since each row $w_r$ appears only in its own $\ell_\infty$‑norm term, the problem can be decomposed. For each $r$ we set
$
b_r \;=\; w_{l,r}\,X_l^T\in\mathbb R^m,
$
and solve independently

$$
\widehat w_{l,r}
\;=\;
\arg\min_{w_r\in Q^d}\;\|\,w_r\,X_l^T - b_r\|_\infty.
$$
In this way, the optimal global quantization of $\ell_\infty$ norms reduces to $n$ parallel regression problems of $\ell_\infty$ norms.

\paragraph{Experiment setup and results}
We apply layer-wise, 3-bit weight quantization to Meta’s OPT-125M model \cite{optmodel}. See Appendix C for full experiment setup details.  At 3-bit precision, we compare AMVM against five baselines, SqueezeLLM \cite{kim2023squeezellm}, GPTQ \cite{frantar2022gptq}, OmniQuant \cite{shao2023omniquant}, ApiQ \cite{liao2024apiq}, and round-to-nearest (RTN). Methods requiring calibration set use the C4 training split, and we measure perplexity on the C4 and WikiText-2 test sets. For all methods, we perform quantization without outlier separation, and after each layer is quantized we forward-propagate its activations. We initialize AMVM via round-to-nearest on an optimized grid and enforce an $L_2$-norm check for every AMVM iteration to ensure the norm does not increase. 
As shown in Table\ref{tab:quantization-results}, AMVM outperforms all baselines on the C4 and WikiText2 datasets, achieving around 14\% improvement on average comparing to these common quantization methods.

\begin{table}[ht!]
\centering
\begin{tabular}{lll}
\hline
Dataset                       
& C4 test split & WikiText2 test split \\ \hline
Original model                         
& 23.41        & 28.32               \\ \hline
RTN  &  3607.45                   
& 5021.97              \\ 
GPTQ                          
& 37.20        & 57.89               \\
SqueezeLLM                    
& 27.86        & 35.62               \\
OmniQuant                     
& 34.58            & 38.26                \\ 
ApiQ                          
& 30.54            & 34.96                \\ 
AMVM(Ours)                          
& \textbf{27.26}        & \textbf{34.70}               \\\hline
\end{tabular}
\caption{Perplexity of the original OPT-125m and its quantized versions with 3-bit weights obtained by six quantization methods on two different datasets. The best perplexity values are shown in bold-face font for each dataset.}
\label{tab:quantization-results}
\end{table}

\paragraph{Impact of GPU acceleration on AMVM}
To isolate the impact of GPU acceleration on worst‑case errors, we quantify the percentage gap between AMVM and Gurobi after 10 seconds. In this experiment, we randomly selected 20 rows from a first‑layer weight matrix of OPT‑125M and ran AMVM at 2‑ and 3‑bit precisions. AMVM was given the same 10s wall‑clock budget on a single CPU core and on a Tesla P100 GPU. Table~\ref{tab:iter_vs_obj} reports the average resulting worst-case residuals ($\ell_\infty$) and average iteration counts. It also reports the average gap to the best objective (UB) found by Gurobi after 7200~s on 12 CPU cores.

The GPU executes 70 iterations versus 12 at 2‑bit (6× faster) and 107 versus 4 at 3‑bit (26× faster), Consequently, the GPU achieves a final gap of $-0.74\%$ at 2‑bit precision and narrows the gap to 0.85\% at 3‑bit precision. Thus, the higher iteration throughput of the GPU directly translates into tighter worst‑case errors under the same time constraint.

\begin{table}[ht]
\centering
\setlength{\tabcolsep}{1mm}
\begin{tabular}{lrrrrrr}
\toprule
& \multicolumn{3}{c}{\textbf{2‑bit}} & \multicolumn{3}{c}{\textbf{3‑bit}}\\
\cmidrule(lr){2-4}\cmidrule(lr){5-7}
\textbf{Platform} & $\ell_\infty$ & Iter. & Gap\% & $\ell_\infty$ & Iter. & Gap\%\\
\midrule
CPU (1‑core) & 2.259 & 12 & 1.23 & 0.990 & 4   & 4.91\\
GPU (P100)   & 2.213 & 70 & $-0.74$ & 0.952 & 107 & 0.85\\
\bottomrule
\end{tabular}
\caption{Objective value ($\ell_\infty$), AMVM iterations completed, and percentage gap to Gurobi’s best objective (UB) after a 10s budget on 20 sampled OPT‑125M rows.}
\label{tab:iter_vs_obj}
\end{table}

\section{Discrete Tomography} \label{sec:tomography}

Our second use case involves discrete tomography, where one seeks to reconstruct an image whose pixel values are restricted to a finite, a priori known set of grey levels  $R = \{\rho_1,\rho_2,\dots,\rho_\ell\}$ \cite{Dart}. Given a non-negative projection matrix  $A \in \mathbb{R}_{\ge0}^{m\times n}$ and measured data $p \in \mathbb{R}^m$, the goal is to find an image $x \;\in\; R^n $ that satisfies the equation $ A\,x \;=\; p$. The reconstruction problem can then be stated as:
$
      \min_{x\in R^n} \;\|A\,x - p\|.
$

In practice, these measurements are corrupted by noise or other errors collectively denoted by $\epsilon$. So, the equation under noise becomes $p = A\,x + \epsilon$ \cite{Dart}. When the noise components $\epsilon_i$ are uniformly distributed, the use of the infinity norm $
\hat x = \arg\min_{x\in R^n}\|A\,x - p\|_\infty
$ is more appropriate than the $\ell_2$ norm. This is due to its connection with the maximum likelihood estimator under uniform noise \cite{Boyd_Vandenberghe_2004,Clason_2012}.

\paragraph{Solution methods.} Several algorithms have been developed for tomographic image reconstruction. One of the most classical algorithms is filtered back projection \cite{fbp}: an analytic method that applies a ramp filter to each projection before back‐projecting to form the image. In contrast, iterative techniques such as the Simultaneous Iterative Reconstruction Technique (SIRT) and the Simultaneous Algebraic Reconstruction Technique (SART) \cite{Sart} repeatedly update all pixel estimates until convergence. While these iterative methods excel for continuous tomography, they do not work well for discrete tomography cases where the object consists of only a few known gray levels. The Discrete Algebraic Reconstruction Technique (DART) \cite{Dart} addresses this by alternating continuous reconstruction steps (e.g.\ using SART) with a discretization step that forces pixel values to the predefined gray levels and classifies pixels as fixed or free for the next iteration. Other approaches, such as stochastic sampling methods \cite{Matej1999, Chan1999} and convex formulations \cite{binarytomo},
 have also been applied to discrete tomography.  Our proposed AMVM algorithm is also applicable for solving discrete tomography.
\paragraph{Experiment setup and results} 
We evaluate AMVM against SIRT, SART, and DART on two tomographic datasets: 
a binary‐image set \cite{binarytomo} and a four‐level (segmented MR/CT low‑grade glioma) set \cite{brainct}. Reconstructions use 64 projection angles with uniform noise (±1000 for binary, ±500 for four‑level). In our DART implementation, SART serves both as the algebraic reconstruction subroutine and as the method for generating the initial solution for AMVM. 
Reconstruction quality is assessed by Mean Absolute Error (MAE) and Structural Similarity Index Measure (SSIM) \cite{SSIM} (see Appendix D for full experiment details).

\begin{table}[ht!]
\centering
\small
\setlength{\tabcolsep}{1mm}
\sisetup{
  detect-weight,   
  mode=text         
}
\begin{tabular}{@{}l l c c c c}
\toprule
& \textbf{Method} & {\textbf{MAE} } & {\textbf{SSIM} } & {\textbf{$\ell_\infty$}} & {\textbf{Time (s)}} \\
\midrule

\multirow{5}{*}{\rotatebox[origin=c]{90}{\textbf{Binary}}} 
& SART       & 12.85$\pm$2.8  & 0.40$\pm$0.07  & 2366 & 0.51    \\
& SIRT       & 5.53$\pm$1.9  & 0.52$\pm$0.03  & 1703 & 25  \\
& DART       & \bfseries 2.87$\pm$1.3  & \bfseries 0.90$\pm$0.03  & 2681 & 68   \\
& AMVM (CPU) & 4.04$\pm$1.3  & 0.74$\pm$0.08  & 1604 & 409  \\
& AMVM (GPU) & 3.82$\pm$1.2  & 0.75$\pm$0.07  & \bfseries 1416 & 61   \\
\midrule
\multirow{5}{*}{\rotatebox[origin=c]{90}{\textbf{Four-level}}} 
& SART       & 16.60$\pm$5.6 & 0.54$\pm$0.08  & 1063 & 0.27    \\
& SIRT       & 14.84$\pm$5.5 & 0.62$\pm$0.1  & 664  & 10   \\
& DART       & 14.70$\pm$6.0 & \bfseries 0.63$\pm$0.1  & 1176 & 128  \\
& AMVM (CPU) & 12.29$\pm$4.7 & 0.58$\pm$0.1  & \bfseries 648  & 1300 \\
& AMVM (GPU) & \bfseries 12.22$\pm$4.6 & 0.57$\pm$0.1  & 707  & 194  \\
\bottomrule
\end{tabular}
\caption{Comparison of reconstruction methods for a binary and a four-level dataset.}
\label{tab:tomography}
\end{table}

The results in Table \ref{tab:tomography} show that AMVM consistently outperforms SART and SIRT methods on both the binary and four‑level datasets in terms of MAE and SSIM. While DART slightly leads in raw MAE and SSIM on the binary dataset, 
on the four-level dataset AMVM reduces MAE by approximately 16\% compared to DART, and it still produces the lowest worst-case $\ell_\infty$-norm) errors for both noise models, making it a robust choice under uniform perturbations. Moreover, the GPU‑accelerated AMVM runs in practical time for both datasets. The CPU‑only variant though nearly matches the accuracy, requires orders of magnitude more computation (especially on the four level images). This underscores that GPU acceleration is essential for large‑scale discrete tomography.

\begin{figure}[ht!]
  \centering
  \makebox[\columnwidth]{%
    \textbf{GT}\hfill
    \textbf{SART}\hfill
    \textbf{SIRT}\hfill
    \textbf{DART}\hfill
    \textbf{AMVM}%
  }\\[4pt]
  
  \makebox[\columnwidth]{%
    \includegraphics[width=0.18\linewidth]{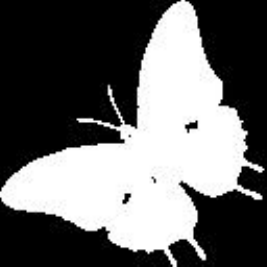}\hfill
    \includegraphics[width=0.18\linewidth]{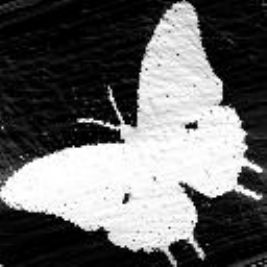}\hfill
    \includegraphics[width=0.18\linewidth]{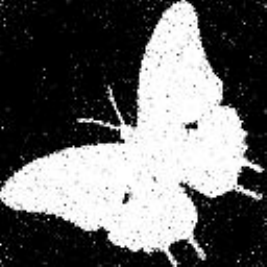}\hfill
    \includegraphics[width=0.18\linewidth]{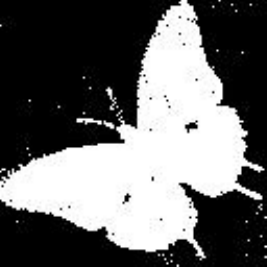}\hfill
    \includegraphics[width=0.18\linewidth]{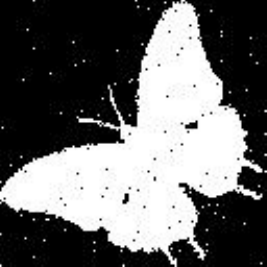}%
  }
  \caption{Reconstruction of a binary butterfly phantom under uniform noise.}
  \label{fig:butterfly}
\end{figure}
In Figure~\ref{fig:butterfly}, we show, from left to right, the Ground Truth (GT) image, and reconstructions produced by SART, SIRT, DART, and AMVM. SART and SIRT employ continuous‑valued updates and therefore produce gray pixels rather than pure black and white. DART enforces binarity and exhibits fewer isolated speckles compared to AMVM. Nevertheless, AMVM delivers a visually comparable binary reconstruction with only a few background speckles. AMVM achieves a substantially lower worst‑case error ($\ell_\infty$) compared to DART, underscoring its robustness against extreme local deviations (consistent with the results in Table~\ref{tab:tomography}).

\section{FIR Design with Discrete Coefficients}\label{sec:FIR}
Digital filters are fundamental components in digital signal processing, used for removing noise from signal and modifying the signal by emphasizing or attenuating specific frequency bands \cite{oppenheim_schafer}. For example, audio equalizers boost the bass or treble in a song. A widely used class of digital filters include the Finite Impulse Response (FIR) filters. An FIR filter of order $N$ produces each output sample $y[n]$ by a weighted sum of the current input sample $x[n]$ and $N$ previous input samples:
$y[n] \;=\;\sum_{k=0}^{N} h_k\,x[n-k].$
Here, the sequence $\{h_k\}$ of length $N+1$ is called the impulse response. FIR filters are always stable, since their output is a finite sum of bounded inputs. They can be designed to have an exactly linear phase, which means they delay all frequency components by the same constant amount~\cite{Rabiner}.
To guarantee a linear phase, a common choice is a symmetric impulse response $ h_k = h_{N-k}\quad (0\le k\le N).$ In this case, the filter’s magnitude response can be written as the cosine polynomial~\cite{oppenheim_schafer}: $H(\omega)\;=\;\sum_{k=0}^{N} h_k\,\cos(k\omega)$.

The design problem is as follows. Given a desired magnitude response $D(\omega)$, find the coefficients $h_k$ that make $H(\omega)$ match $D(\omega)$ as closely as possible.   For example, for a low-pass filter, the desired response is typically defined to pass low frequencies and block high frequencies using
$$
D(\omega)=
\begin{cases}
K, & 0\le\omega\le\omega_p,\\[6pt]
0, & \omega_s\le\omega\le\pi.
\end{cases}
$$
In this equation, $K$ is the passband gain (often set to 1), $\omega_p$ is the passband edge, $\omega_s$ is the stopband edge, and the interval $\omega_p<\omega<\omega_s$ is the transition band. Among FIR methods, the equiripple (Chebyshev) design directly minimizes the maximum deviation (ripple)  defined by
\[
\delta = \max_{\omega\in[0,\omega_p]\cup[\omega_s,\pi]} \bigl|H(\omega)-D(\omega)\bigr|,
\]
and finds the filter coefficients $h$ by solving
$
\min_{h}\,\delta.
$
Thus, it minimizes the maximum deviation between the actual response $H(\omega)$ and the desired response $D(\omega)$ over both the passband and stopband \cite{remez, Fir}.

This problem can be solved for continuous coefficients using the Remez exchange algorithm \cite{remez}, which iteratively adjusts extremal frequencies to equalize ripple magnitude and converges to the optimal minimax solution. However, practical hardware implementations require the filter coefficients to be represented with a finite word-length \cite{Fir, CHANDRA2016212}. Several applicable schemes exist, such as the signed power of two format \cite{SP2} and the fixed-point representation \cite{Fir, Hung}. We focus on the latter which is more relevant to our topic. For a precision of $p$ bits, this scheme constrains each coefficient to the discrete set $G = \left\{ {k}/{2^{p-1}} \;\middle|\; k \in \mathbb{Z}, -2^{p-1} \le k \le 2^{p-1}-1 \right\}$.
Additionally, to formulate the problem for a numerical solution, the continuous frequency domain is discretized into a dense grid of $m$ points ($\omega_1, \dots, \omega_m$).

The filter design task under fixed-point representation \cite{Fir, Hung} boils down to the mathematical task of finding the coefficient vector $x$ that solves the following minimax optimization problem:
\[
\min_{x \in G^{n}} \max_{j=1,\dots,m} |(Ax - b)_j|.
\]
Here, $x = [h_0, \dots, h_N]^T$ is the vector of $n=N+1$ coefficients, the matrix $A$ is defined by $A_{jk} = \cos(k\omega_j)$, and the vector $b$ is the sampled desired response, $b_j = D(\omega_j)$.

Common exact solutions formulate the minimax design as a mixed integer linear program, mixed integer semi-infinte linear program \cite{Fir}, or semi definite programming \cite{dsp_fir}, using auxiliary variables to linearize the peak ripple and solve it by branch and bound. A faster heuristic solution involves computing the continuous equiripple coefficients using the Remez exchange algorithm and then rounding each tap to the nearest discrete level \cite{remez, Fir, CHANDRA2016212}. A third approach is based on a simple local‐search procedure. It starts from the rounded solution of the Remez exchange algorithm. It then iteratively perturbs a single coefficient to one of its nearest adjacent levels ($h\pm1$) accepting the change only if it improves the objective \cite{LS_fir}. Our AMVM heuristic is also applicable for solving this minimax design problem.

\paragraph{Experiment Setup}
We compare our AMVM method to two baselines for the FIR filter design task: the optimal integer programming solution obtained by Gurobi and the standard approach of rounding the coefficients produced by the Remez exchange algorithm  (rounded for short) \cite{remez}. We use the continuous Remez exchange solution as our initial solution.

\paragraph{FIR Experiment 1}
Our experiment 1A follows exactly the setup in \cite{Fir}: the passband and stopband frequencies are $ \omega_p = 2\pi/5,  \omega_s = 4\pi/7,$
with scaling factor \(K=1\) and coefficient bit-depth \(p=4\) and $N = 12$. For experiment 1B, we chose a much narrower transition band between $\omega_p = 2\pi/200$ and $\omega_s = 3\pi / 200$ to test sharper frequency selectivity. This tighter bandwidth necessitates a higher filter order ($N=500$) to meet the attenuation requirements, and we increased the coefficient bit-depth to $p=8$ to keep quantization error under control at this higher order.

For the relatively small instance in experiment 1A, our method matches Gurobi’s optimal solution (as shown in Table~\ref{tab:fir_results}). In experiment 1B, Gurobi did not find any feasible solution within a 10s time limit. Even when the limit was increased to 30s, it only achieved a MIP gap of 83.86 \%. By contrast, our algorithm produced a better solution in only 5s (see Table~\ref{tab:fir_results}). These results demonstrate the practical applicability of our AMVM heuristic in the context of FIR filter design.

\begin{table}[ht!]
\centering
\begin{tabular}{p{2cm}cccc}
\toprule
\multirow{2}{*}{\textbf{Method}}
  & \multicolumn{2}{c}{ ($N=12,p=4$)} 
  & \multicolumn{2}{c}{ ($N=500,p=8$)} \\
\cmidrule(lr){2-3} \cmidrule(lr){4-5}
  & $\delta$ & Time (s)  & $\delta$ & Time (s)  \\
\midrule
Rounded & 0.264 & 0.0003 & 0.31 & 0.0084 \\
Gurobi        & \textbf{0.207} & 0.04   & 0.26 & 30.0   \\
AMVM(CPU)   & 0.207   & 2.8    &0.13   &   5.5  \\
AMVM(GPU)   & \textbf{0.207} & 2.0    & \textbf{0.13} & 5.0    \\
\bottomrule
\end{tabular}
\caption{Performance measures of AMVM and two other methods for FIR filter design in experiments 1A ($N=12,p=4$) and 1B ($N=500,p=8$).}
\label{tab:fir_results}
\end{table}

\paragraph{FIR Experiment 2}
This experiment involves an anti-Hum filter for 60 Hz power-line rejection. In countries with a 60 Hz AC mains supply, a narrow stopband filter is used to deeply attenuate the hum at 60 Hz while passing all other frequencies \cite{bdstop}. To evaluate our method on a demanding, high-fidelity application, we configure the filter with a high order of $N=500$ and a coefficient bit-depth of $p=8$. In its ideal form, the desired frequency response $D(\omega)$ is defined piece-wise, $D(\omega)=0$ if $\omega_{p1}\le\omega\le\omega_{p2}$ and $D(\omega)=1 $ otherwise.

where the band-edge frequencies are $(\omega_{s1},\,\omega_{p1},\,\omega_{p2},\,\omega_{s2})
=2\pi\!\left({58} / {F_s},\;{59} / {F_s},\;{61} / {F_s},\;{62} / {F_s}\right)$ with $F_s = 1\text{\,kHz}.$ Table~\ref{tab:fir_results_AntiHum} compares different methods on solution quality (ripple $\delta$, where lower is better) and computation time.

\begin{table}[ht!]
\centering
\begin{tabular}{lcc}
\toprule
\textbf{Method} & \(\delta\) & Time (s) \\
\midrule
Rounded           & 0.39            & 0.006  \\
Gurobi            & 0.24            & 120.0    \\
AMVM (CPU)        & 0.20            & 20.0     \\
AMVM (GPU)        & \textbf{0.19}   & 20.0    \\
\bottomrule
\end{tabular}
\caption{Performance Comparison for Anti-Hum Filter}
\label{tab:fir_results_AntiHum}
\end{table}
The results in Table~\ref{tab:fir_results_AntiHum} are significant. Our GPU-accelerated AMVM finds a higher-quality solution ($\delta=0.19$) than the state-of-the-art Gurobi solver ($\delta=0.24$) in just 20 seconds, making it six times faster. This demonstrates that our proposed method is a highly efficient  for the FIR design problem.
\section{Discussion and Future Work}
We introduced the DMMV problem as a fundamental optimization problem for the generic application of worst‐case error optimization tasks. We then proposed AMVM, a GPU‐accelerated heuristic based on adaptive large neighborhood search, and demonstrated its applicability on three use cases: post‐training quantization of LLMs (Table~\ref{tab:quantization-results}), discrete tomography (Table~\ref{tab:tomography}), and FIR filter design (Tables~\ref{tab:fir_results} and \ref{tab:fir_results_AntiHum}). Our comparative assessments against use-case-specific methods demonstrate that a general‐purpose DMMV solver can match or exceed the performance of specialized methods. In particular, AMVM achieves on average 14\% perplexity improvement over existing LLM quantizers, reduces worst‐case reconstruction error by 16\% under uniform noise, and attains nearly 50\% ripple reduction compared to a commercial MILP solver. These results illustrate that even without problem‐specific customization, a context‐free DMMV approach can deliver superior performance. The technical details of AMVM are provided in the Materials and Methods Section.

\paragraph{Limitations and Future Work} While AMVM offers promising empirical performance, it remains a heuristic without optimality guarantees. Its effectiveness partially depends on a suitable initial solution and on hyperparameter choices (e.g., destroy/repair scores, $\alpha$ in Eq.4 (see Appendix B). 
 Furthermore, our study focuses on three applications; extensions to other settings (e.g., robust regret minimization) remain open avenues for further explorations. Future research includes extending DMMV to new discrete‐design domains such as robust scheduling under failure scenarios\cite{Jensen2001}, and robust regret minimization\cite{Poursoltani} to better assess the extent of its applicability. Furthermore, one may use our open-source research code and software to develop a scalable multi‐GPU and distributed versions of AMVM, optimizing data movement and kernel fusion (e.g., via CUDA‐MPI/NCCL) \cite{nccl} to tackle larger‐scale or real‐time DMMV instances.

\section{Materials and Methods}\label{sec:algorithm}

In this section, we provide the technical details of AMVM (Algorithm ~\ref{alg:AMVM}). We start by describing the destruction and repair operators, followed by explaining the local search. We specifically define the necessary condition for solution improvement and use these conditions as acceleration filters. Finally, we explain how the implementation has been adapted for GPU-based acceleration.

\begin{algorithm}[htbp!]
\caption{GPU-accelerated AMVM framework}
\label{alg:AMVM}
\begin{algorithmic}[1]
\Require Initial solution $x_0 \in \mathbb{V}^n$
\Ensure Best solution $x^*$ minimizing $\|Ax - b\|_\infty$
\State $x^* \gets x_0$
\While{stopping criterion not met}
    \State Select destroy and repair operators $(d, r)$ according to current weights
    \State $x_t^\text{partial} \gets d(x_t)$               \Comment{Apply destroy operator}
    \State $\hat{x}_t \gets r(x_t^\text{partial})$             \Comment{Apply repair operator}
    \State $x_{t+1} \gets \mathrm{LS}(\hat{x}_t )$   \Comment{Improve via local search}
    \If{acceptance criterion holds for $x_{t+1}$}
        \State $x_t \gets x_{t+1}$
        \If{$\|A x_{t+1} - b\|_\infty < \|A x^* - b\|_\infty$}
            \State $x^* \gets x_{t+1}$        \Comment{Update best solution}
        \EndIf
        \State Update operator‐selection weights
    \EndIf
\EndWhile
\State \Return $x^*$
\end{algorithmic}
\end{algorithm}
\paragraph{Search Operators}
At each iteration, AMVM applies one of two destroy operators and then one of two repair operators (see Appendix B for full details). In brief:

\begin{itemize}
  \item \textbf{Random Destroy / Repair.}  
    \emph{Destroy:} pick a small set of variables at random and unassign them.  
    \emph{Repair:} for each unassigned variable, pick one of its two nearest values from \(\mathbb V\) at random.
  \item \textbf{Worst-Remove Destroy.}  
    Compute an impact score \(d_j\) for each variable \(x_j\) (see Appendix B), then choose \(r\) different variables with probability
    $
      P_j \;=\;{d_j}/ {\sum_{i=1}^n d_i}
    $
    and unassign them.
  \item \textbf{Greedy Repair.}  
    For each unassigned variable, try its two nearest values and keep the one that gives the smallest worst-case error.
\end{itemize}

After repair, we immediately apply two local-search moves: (1) a \textbf{1-OPT} move that shifts each variable to its nearest neighbor if that improves the error, and (2) a \textbf{swap} (see Algorithm 2 in Appendix B) that exchanges two variable values only when it reduces the worst-case error. While the exact condition for such an improvement can be defined, a brute-force search is computationally prohibitive. We uses acceleration techniques based on the following mathematical property of DMMV to find a improving swaps efficiently.

For a given solution $x$ we define the residual vector as $s =  Ax -b$, and the objective value as $t = \|s\|_\infty$. For a potential swap between $x_i$ and $x_j$ (with $x_i > x_j$), the difference is defined as $\Delta = x_i - x_j >0 $
\begin{proposition}[Strict‐Improvement Condition via Swap] \label{prop:exact_condition}
A swap between elements $x_i$ and $x_j$ produces a strictly improved solution with a new objective value $t' < t$ iff the following condition holds:
\begin{equation} \label{eq:prop1_condition}
  \frac{-t - s_k}{\Delta} < a_{kj} - a_{ki} < \frac{t - s_k}{\Delta} \quad  \forall k=1,\dots,m.
\end{equation}
\end{proposition}
The proof of this proposition is given in Appendix A.
It is computationally expensive to verify the condition in Proposition \ref{prop:exact_condition} for all possible swaps directly. Therefore, our heuristic in Algorithm 2 (appendix B) uses following filters to find a good swap without performing an exhaustive search. 

\begin{corollary}[Swap Candidate Selection] \label{lem:swap_selection}
A list of promising candidate swaps can be generated efficiently by checking a necessary improvement condition against only a small subset of rows. Instead of checking all $m$ rows, we select the $k_\varepsilon$ rows with the largest residuals $|s_k|$. A pair $(i,j)$ is selected as a candidate if it satisfies the condition 
\[
\begin{cases}
    a_{kj} - a_{ki} < \epsilon_k /\Delta & \text{if } s_k > 0 \\
    a_{kj} - a_{ki} > -\epsilon_k /\Delta & \text{if } s_k < 0.
\end{cases}
\]
In here, $\epsilon_k = t - |s_k|$ for every row $k$ in this $k_\varepsilon$-subset. This serves as the first filtering stage of the algorithm.
\end{corollary}
\begin{corollary}[Row Screening Test] \label{lem:row_screening}
For a given swap operation, a row $k$ is guaranteed to satisfy the improvement condition in Proposition \ref{prop:exact_condition} if it satisfy the following condition:
\[
\frac{-t - s_k}{\Delta} < A_k^- \quad\text{and}\quad A_k^+ < \frac{t - s_k}{\Delta},
\]
where $
A_k^- \;=\;\min_{i,j}\bigl(a_{kj}-a_{ki}\bigr)$  $ A_k^+ \;=\;\max_{i,j}\bigl(a_{kj}-a_{ki}\bigr)$

\end{corollary}

\paragraph{GPU Acceleration}
To fully exploit modern hardware, we implement every phase of our heuristic destruction, repair, local search and swap filtering/evaluation as fused CUDA kernels within the PyTorch framework. All key data structures (the constraint matrix $A$, solution vector $x$, residuals $s = Ax - b$, bound tensors, distance measures, and impact scores) reside on the GPU as dense \texttt{torch.Tensor} objects. 
Residual and objective computations, namely the matrix–vector product $s=Ax-b$ and the $\ell_\infty$ norm $\|s\|_\infty$, are executed in parallel via batched linear algebra and reduction primitives. Destroy‐phase scoring leverages broadcasted tensor arithmetic and element‐wise exponentials to compute bound distances $R_j^k$ and impact scores $d_j$. Then, removals are sampled through a parallel multinomial draw. Repair‐phase moves are similarly vectorized: greedy repairs evaluate all candidates in bulk and choose minima via global reductions.

Swap filtering constructs a three‐dimensional difference tensor between candidate levels and applies the necessary conditions of Proposition~\ref{prop:exact_condition} and Corollary~\ref{lem:swap_selection} as masked comparisons, extracting promising $(i,j)$ pairs without explicit loops. These candidates are then evaluated in batched groups, computing both $\ell_\infty$ and $\ell_2$ objective changes in parallel and selecting the best swap by efficient tensor‐wide argmin operations. We observe a significant speedup (i.e., up to $26\times$) compared to a highly optimized single-CPU implementation by fusing these operations into a handful of GPU kernels (to avoid Python‐level iteration and CPU–GPU synchronization) as much as possible.

 \section*{Acknowledgments}
 Authors acknowledge Bill Lin for help with algorithm development, Ali Hadizadeh for the helpful discussions and Yixin Yin for technical assistance.

\bibliography{aaai2026}
\clearpage
\input{Appendix1}
\end{document}

%% file: Appendix1.tex
\section{Appendix A: Proof of Proposition 1}
\begin{proof}
Using the standard basis vectors $e_i$ and $e_j$, we can write the new solution vector $x'$ as:
\[
x' = x + (x_j - x_i)e_i + (x_i - x_j)e_j
\]
Let's define a scalar value $\Delta = x_i - x_j$. This represents the difference in the values that are swapped. Note that $x_j - x_i = -\Delta$. Substituting this gives:
\[
x' = x - \Delta e_i + \Delta e_j = x + \Delta(e_j - e_i)
\]

Now, we derive the new residual vector, $s'$ is :
\begin{align*}
    s' &= Ax' - b                                     \\
       &= A(x + \Delta(e_j - e_i)) - b               \\
       &= Ax - b + \Delta(Ae_j - Ae_i)                
\end{align*}
Since $Ae_i = a_i$ and $Ae_j = a_j$ (the $i$-th and $j$-th columns of $A$), we arrive at the final expression for the new residual:
\[
\boxed{s' = s + (x_i - x_j)(a_j - a_i)}
\]
In component form, for each row $k = 1, \dots, m$, the new residual component $s'_k$ is:
\[
s'_k = s_k + (x_i - x_j)(a_{kj} - a_{ki})
\]

A swap move results in a strict improvement if and only if the new objective value $t' = \norm{s'}_\infty$ is strictly less than the old objective value $t$.
\[
\norm{s + (x_i - x_j)(a_j - a_i)}_\infty < t
\]
This is equivalent to the system of $m$ simultaneous inequalities:
\[
|s_k + (x_i - x_j)(a_{kj} - a_{ki})| < t \quad \text{for all } k \in \{1, \dots, m\}
\]
The condition is $|s_k + \Delta \cdot  (a_{kj} - a_{ki})  | < t$.

For a given swap $(i, j)$. This move leads to an improvement if $\Delta$ satisfies the required condition for all $m$ rows. The condition for each row $k$ is:
\[
\boxed{-t - s_k < \Delta \cdot (a_{kj} - a_{ki}) < t - s_k}
\]

Since $\Delta = x_i - x_j > 0$,  \[ \boxed{\frac{-t - s_k}{\Delta} < a_{kj} - a_{ki} < \frac{t - s_k}{\Delta}} \] This condition must hold for all $k=1,\dots,m$ for the swap to be a strict improvement, which completes the proof of Proposition 1. 
\end{proof}

\section{Appendix B: Operator Details and Swap Move}
\label{app:operators}

\subsection{B.1 Random Destroy / Repair}
\begin{itemize}
  \item \textbf{Random Destroy.}  
    Select \(r\) distinct variables at random and unassign them.
  \item \textbf{Random Repair.}  
    For each unassigned variable, choose one of its two nearest values in \(\mathbb V\) at random.
\end{itemize}

\subsection{B.2 Worst-Remove Destroy}
Given current solution \(x\):
\begin{enumerate}
  \item Compute residuals \(s = A\,x - b\) and worst-case error \(t = \|s\|_\infty\).
  \item For each variable \(j\) and each constraint \(k\), set
    \begin{equation}
      LB_j^k =
       \frac{-t + b_k - \sum_{i\neq j}a_{ki}x_i}{a_{kj}},
      \end{equation}
      \begin{equation}
      UB_j^k = \frac{\,t + b_k - \sum_{i\neq j}a_{ki}x_i}{a_{kj}},
    \end{equation}
    \begin{equation}
      R_j^k = \min\{\,UB_j^k - x_j,\;x_j - LB_j^k\}.
    \end{equation}
  \item Compute impact scores
  \begin{equation}
    \label{eq:impact}
d_j = \frac{\sum_{k=1}^m |s_k| e^{-\alpha R_j^k}}{\sum_{k=1}^m |s_k|}.    
\end{equation}
  \item Pick \(r\) different variables by sampling from \(\{1,\dots,n\}\) with probabilities
    \(\;P_j = d_j / \sum_{i=1}^n d_i.\)
\end{enumerate}

\subsection{B.3 Greedy Repair}
For each unassigned variable \(j\):
\begin{enumerate}
  \item Let \(v^-\) and \(v^+\) be the two values in \(\mathbb V\) closest to the old \(x_j\).
  \item Form two candidate solutions by setting \(x_j = v^-\) or \(v^+\).
  \item Compute their worst-case errors \(\|A x' - b\|_\infty\) and choose the smaller.
\end{enumerate}

\subsection{Swap Move}
We look for a pair \((i,j)\) with \(x_i>x_j\) whose exchange lowers \(\|A x - b\|_\infty\). For each \((i,j)\) with \(\Delta = x_i - x_j\),  compute the new residual \(s' = s + \Delta\,(a_j - a_i)\) and error \(t'=\|s'\|_\infty\).  Keep the swap with smallest \(t'\).
\setcounter{algorithm}{1}
\begin{algorithm}[ht!]
\caption{Swap Algorithm with Filtering}\label{alg:swap}
\begin{algorithmic}[1]
\Require Problem data $A, b$, solution $x$
\Ensure Best swap $(i^*, j^*)$, or None

\State $s \gets Ax - b$
\State $t \gets \|s\|_\infty$
\State $(i^*, j^*) \gets \text{None}$
\State $t_{best} \gets t$

\Statex
\State \textbf{Step 1: Find Promising Swaps}
\State \Comment{Use corollary 1 for candidate generation.}
\State $C \gets$ FindCandidates($x, s, t, A$)

\Statex
\State \textbf{Step 2: Evaluate and Select Best}
\ForAll {candidate $(i, j) \in C$}
    \State \Comment{Use corollary 2 to accelerate check.}
    \If {IsImproving($x, s, t, A, (i,j)$)}
        \State $s' \gets s + (x_i - x_j)(a_j - a_i)$
        \State $t' \gets \|s'\|_\infty$
        \If {$t' < t_{best}$}
            \State $t_{best} \gets t'$
            \State $(i^*, j^*) \gets (i, j)$
        \EndIf
    \EndIf
\EndFor
\State \Return $(i^*, j^*)$
\end{algorithmic}
\end{algorithm}

\section*{Appendix C: Quantization Experiment Details}

\subsection*{C.1 Hardware and Parallelism}
\begin{itemize}
  \item \textbf{GPUs:} 4$\times$NVIDIA Tesla P100 (16GB each).
  \item \textbf{Concurrency:} 10 quantization rows processed per GPU in parallel (40 rows total).
  \item \textbf{Per‐row runtime:} AMVM runs for 10seconds per row.
\end{itemize}

\section*{Appendix D: Experimental Details for Discrete Tomography }

\subsection*{D.1 Reconstruction parameters}
\begin{itemize}
  \item \textbf{AMVM:} 10 total iterations.
  \item \textbf{SIRT:} 1000 iterations.
  \item \textbf{SART:} 1000 iterations.
  \item \textbf{DART:} 100 iterations of DART, each using SART as the sub‑routine (1000 SART iterations per DART iteration).
\end{itemize}

\subsection*{D.2 Dataset preparation}
\begin{itemize}
  \item \textbf{Binary‑image set} \cite{binarytomo}:
    \begin{itemize}
      \item 22 images, scaled to $128 \times 128$ pixels.
    \end{itemize}
  \item \textbf{Four‑level MR/CT glioma set} \cite{brainct}:
    \begin{itemize}
      \item 22 Images scaled to $128 \times 128$ pixels.
      \item Segmented into four gray‑levels via uniform 2‑bit quantization (levels 0, 85, 170, 255).
    \end{itemize}
\end{itemize}

\subsection*{D.3 Implementation}
\begin{itemize}
  \item Algorithm implementations (DART, SIRT, SART) from the DART\_python repository:
    \url{https://github.com/OhGreat/DART_python/tree/main}
  \item ASTRA toolbox used for projection-matrix construction and sinogram generation \cite{astra1, astra2}.
   \item GPU: One NVIDIA Tesla P100 with 16GB of memory.
\end{itemize}

\subsection*{D.4 Projection geometry}
\begin{itemize}
  \item \textbf{Projections:} 64 angles, uniformly spaced from $0$ to $\pi$.
  \item \textbf{Beam geometry:} Parallel‑beam.
  \item \textbf{Detector:} 128 pixels with 1‑pixel spacing.
\end{itemize}

\section*{Appendix E: FIR Design Experiment Details}

\subsection*{E.1 : FIR Experiment 3}
\begin{figure}[htp!]
  \centering
\includegraphics[width=0.45\textwidth]{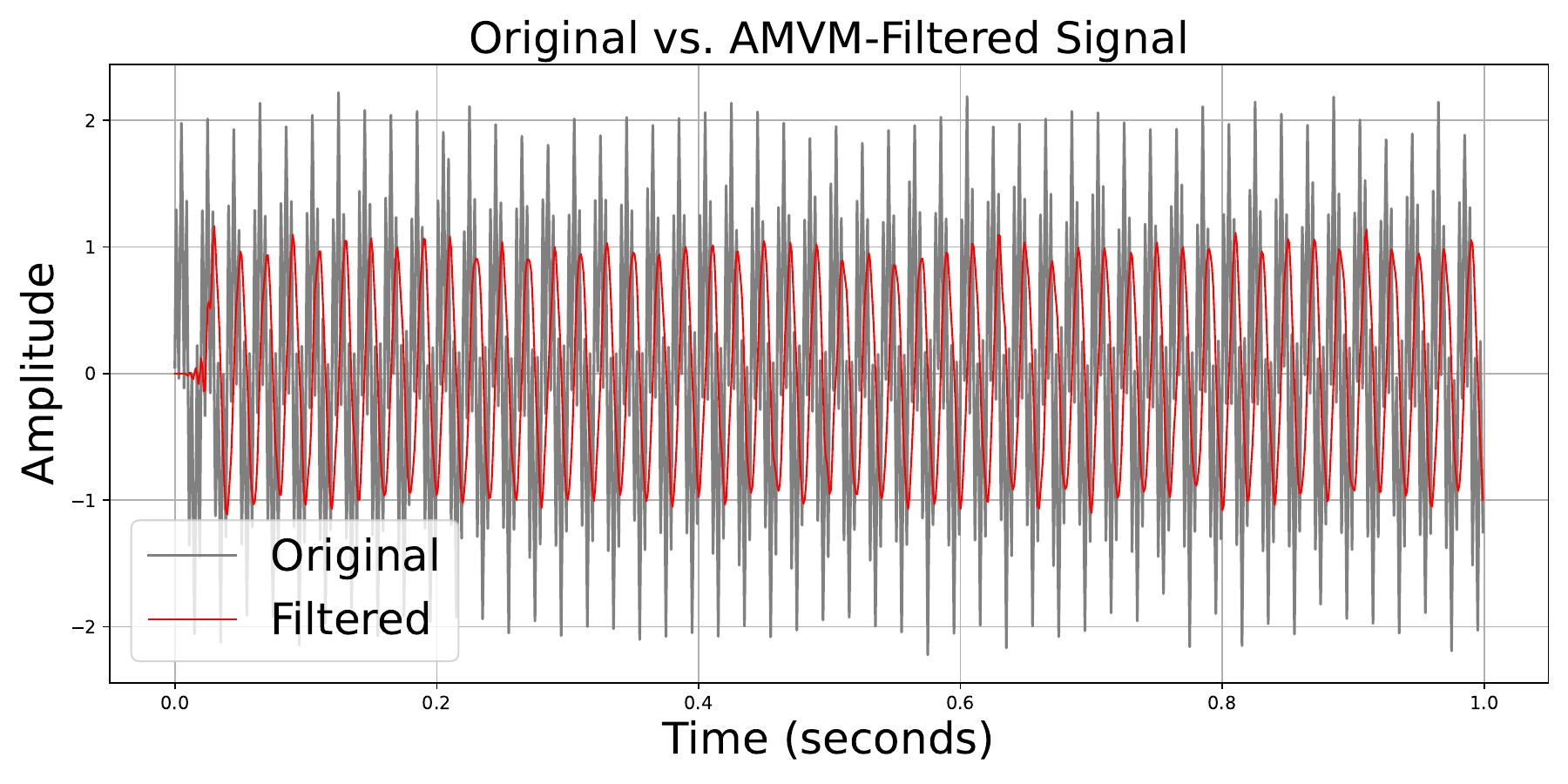}
    \caption{FIR low-pass filtering with 200 Hz cutoff}
    \label{fig:2tone}
\end{figure}

This experiment involves the task of two tone low‑pass FIR filtering. We generated a 1s, 1kHz‑sampled two‑tone signal composed of 50 Hz and 250 Hz sinusoids and added white Gaussian noise ($\sigma^2=0.01$). Using our ALNS algorithm in 8‑bit precision, we designed and applied a 51‑tap linear‑phase FIR low‑pass filter with a 200Hz cutoff. Figure\ref{fig:2tone} clearly shows the 250 Hz component and noise being attenuated, while the 50 Hz tone remains intact.

\subsection*{E.2 Experimental setup}
\begin{itemize}
  \item \textbf{Methods:} AMVM vs.\ optimal integer programming (Gurobi) and coefficient‐rounding of the Remez exchange solution.
  \item \textbf{Initial solution:} Continuous Remez exchange output.
  \item \textbf{Grid size:} $16N$.
  \item \textbf{GPU:} One NVIDIA Tesla P100 with 16\,GB memory.
  \item \textbf{Iterations:} 100 iterations for AMVM.
\end{itemize}

\section*{Appendix F: AMVM Algorithm Hyperparameters}

\subsection*{F.1 AMVM Algorithm Settings}
\begin{itemize}
  \item \textbf{$k_{\epsilon}$-subset size:} 100.
  \item \textbf{Destroy rate:} 0.5\% of the $x_i$ variables.
  \item \(\alpha\) (worst‐removal operator coefficient): 0.3.
\end{itemize}
\subsection*{F.2 AMVM Operator Selections}
AMVM employs adaptive roulette-wheel\cite{ALNS} selection to choose among the four operators. Each operator \(i\) is assigned a nonnegative weight \(w_i\) based on its recent success, and these weights are normalized into probabilities. An operator is then sampled according to \(\{p_i\}\), favoring high‐performing moves while still allowing less successful ones to be selected, thus balancing exploitation and exploration.

%% file: main.bbl
\begin{thebibliography}{51}
\providecommand{\natexlab}[1]{#1}

\bibitem[{Andersen and Kak(1984)}]{Sart}
Andersen, A.; and Kak, A. 1984.
\newblock Simultaneous Algebraic Reconstruction Technique ({SART}): A superior implementation of the ART algorithm.
\newblock \emph{Ultrasonic Imaging}, 6(1): 81--94.

\bibitem[{Barrodale and Phillips(1975)}]{Barrodale}
Barrodale, I.; and Phillips, C. 1975.
\newblock Algorithm 495: Solution of an Overdetermined System of Linear Equations in the Chebychev Norm [F4].
\newblock \emph{ACM Trans. Math. Softw.}, 1(3): 264–270.

\bibitem[{Batenburg and Sijbers(2011)}]{Dart}
Batenburg, K.~J.; and Sijbers, J. 2011.
\newblock {DART}: A Practical Reconstruction Algorithm for Discrete Tomography.
\newblock \emph{IEEE Transactions on Image Processing}, 20(9): 2542--2553.

\bibitem[{Boyd and Vandenberghe(2004)}]{Boyd_Vandenberghe_2004}
Boyd, S.; and Vandenberghe, L. 2004.
\newblock \emph{Convex Optimization}.
\newblock Cambridge University Press.

\bibitem[{Cao and Aref(2025)}]{cao2025saliency}
Cao, D.; and Aref, S. 2025.
\newblock Enhancing Ultra-Low-Bit Quantization of Large Language Models Through Saliency-Aware Partial Retraining.
\newblock In \emph{Proceedings of the 22nd International Conference on Modeling Decisions for Artificial Intelligence}, 354--383.
\newblock LNAI 15957.

\bibitem[{Chan, Herman, and Levitan(1999)}]{Chan1999}
Chan, M.~T.; Herman, G.~T.; and Levitan, E. 1999.
\newblock \emph{Probabilistic Modeling of Discrete Images}, 213--235.
\newblock Boston, MA: Birkh{\"a}user Boston.
\newblock ISBN 978-1-4612-1568-4.

\bibitem[{Chandra and Chattopadhyay(2016)}]{CHANDRA2016212}
Chandra, A.; and Chattopadhyay, S. 2016.
\newblock Design of hardware efficient FIR filter: A review of the state-of-the-art approaches.
\newblock \emph{Engineering Science and Technology, an International Journal}, 19(1): 212--226.

\bibitem[{Clason(2012)}]{Clason_2012}
Clason, C. 2012.
\newblock $L_\infty$ fitting for inverse problems with uniform noise.
\newblock \emph{Inverse Problems}, 28(10): 104007.

\bibitem[{Devlin et~al.(2019)Devlin, Chang, Lee, and Toutanova}]{BERT}
Devlin, J.; Chang, M.-W.; Lee, K.; and Toutanova, K. 2019.
\newblock {BERT}: Pre-training of Deep Bidirectional Transformers for Language Understanding.
\newblock In \emph{Proceedings of the 2019 conference of the North American chapter of the association for computational linguistics: human language technologies, volume 1 (long and short papers)}, 4171--4186.

\bibitem[{Frantar et~al.(2022)Frantar, Ashkboos, Hoefler, and Alistarh}]{frantar2022gptq}
Frantar, E.; Ashkboos, S.; Hoefler, T.; and Alistarh, D. 2022.
\newblock {GPTQ}: Accurate post-training quantization for generative pre-trained transformers.
\newblock \emph{arXiv preprint arXiv:2210.17323}.

\bibitem[{Gillis and Shitov(2019)}]{GILLIS2019367}
Gillis, N.; and Shitov, Y. 2019.
\newblock Low-rank matrix approximation in the infinity norm.
\newblock \emph{Linear Algebra and its Applications}, 581: 367--382.

\bibitem[{{Gurobi Optimization LLC}(2025)}]{gurobi}
{Gurobi Optimization LLC}. 2025.
\newblock {Gurobi Optimizer Reference Manual}.
\newblock \url{https://docs.gurobi.com/projects/optimizer/en/current/}.

\bibitem[{Huang et~al.(2024)Huang, Liu, Qin, Li, Zhang, Liu, Magno, and Qi}]{huang2024billm}
Huang, W.; Liu, Y.; Qin, H.; Li, Y.; Zhang, S.; Liu, X.; Magno, M.; and Qi, X. 2024.
\newblock Bi{LLM}: Pushing the limit of post-training quantization for {LLM}s.
\newblock In \emph{Proceedings of the 41st International Conference on Machine Learning}, 20023--20042.

\bibitem[{Ito and Hirabayashi(2006)}]{SP2}
Ito, R.; and Hirabayashi, R. 2006.
\newblock Optimal design of {FIR} filter with SP2 coefficients based on semi-infinite linear programming method.
\newblock In \emph{2006 14th European Signal Processing Conference}, 1--5.

\bibitem[{Ito, Suyama, and Hirabayashi(2001)}]{Fir}
Ito, R.; Suyama, K.; and Hirabayashi, R. 2001.
\newblock Optimal design of {FIR} filter with discrete coefficients based on integer semi-infinite linear programs.
\newblock In \emph{ISCAS 2001. The 2001 IEEE International Symposium on Circuits and Systems}, volume~2, 629--632 vol. 2.

\bibitem[{Jensen(2001)}]{Jensen2001}
Jensen, M.~T. 2001.
\newblock Finding worst-case flexible schedules using coevolution.
\newblock In \emph{Proceedings of the 3rd Annual Conference on Genetic and Evolutionary Computation}, GECCO'01, 1144–1151. San Francisco, CA, USA: Morgan Kaufmann Publishers Inc.
\newblock ISBN 1558607749.

\bibitem[{Jin et~al.(2024)Jin, Du, Huang, Liu, Luan, Wang, and Xiong}]{jin2024comprehensive}
Jin, R.; Du, J.; Huang, W.; Liu, W.; Luan, J.; Wang, B.; and Xiong, D. 2024.
\newblock A comprehensive evaluation of quantization strategies for large language models.
\newblock In \emph{Findings of the Association for Computational Linguistics ACL 2024}, 12186--12215.

\bibitem[{{Kadu} and {van Leeuwen}(2020)}]{binarytomo}
{Kadu}, A.; and {van Leeuwen}, T. 2020.
\newblock A Convex Formulation for Binary Tomography.
\newblock \emph{IEEE Transactions on Computational Imaging}, 6: 1--11.

\bibitem[{Kahl(2005)}]{Kahl2005}
Kahl, F. 2005.
\newblock Multiple view geometry and the L/sub /spl infin//-norm.
\newblock In \emph{Tenth IEEE International Conference on Computer Vision (ICCV'05) Volume 1}, volume~2, 1002--1009 Vol. 2.

\bibitem[{Kiambi, Mwangi, and Kamucha(2022)}]{PARR}
Kiambi, S.; Mwangi, E.; and Kamucha, G. 2022.
\newblock Reducing PAPR of OFDM signals using a tone reservation method based on $\ell_{\infty}$-norm minimization.
\newblock \emph{Journal of Electrical Systems and Information Technology}, 9.

\bibitem[{Kim et~al.(2024)Kim, Hooper, Gholami, Dong, Li, Shen, Mahoney, and Keutzer}]{kim2023squeezellm}
Kim, S.; Hooper, C.; Gholami, A.; Dong, Z.; Li, X.; Shen, S.; Mahoney, M.~W.; and Keutzer, K. 2024.
\newblock SqueezeLLM: dense-and-sparse quantization.
\newblock In \emph{Proceedings of the 41st International Conference on Machine Learning}, 23901--23923.

\bibitem[{Kleinberg and Tardos(2005)}]{subsetsum}
Kleinberg, J.; and Tardos, E. 2005.
\newblock \emph{Algorithm Design}.
\newblock USA: Addison-Wesley Longman Publishing Co., Inc.
\newblock ISBN 0321295358.

\bibitem[{Kodek and Steiglitz(1981)}]{LS_fir}
Kodek, D.; and Steiglitz, K. 1981.
\newblock Comparison of optimal and local search methods for designing finite wordlength FIR digital filters.
\newblock \emph{IEEE Transactions on Circuits and Systems}, 28(1): 28--32.

\bibitem[{Koo et~al.(2024)Koo, Park, Jung, and Kung}]{koo2024opal}
Koo, J.; Park, D.; Jung, S.; and Kung, J. 2024.
\newblock {OPAL}: Outlier-Preserved Microscaling Quantization Accelerator for Generative Large Language Models.
\newblock In \emph{Proceedings of the 61st ACM/IEEE Design Automation Conference}, 1--6.

\bibitem[{Lang, Guo, and Huang(2024)}]{lang2024comprehensive}
Lang, J.; Guo, Z.; and Huang, S. 2024.
\newblock A comprehensive study on quantization techniques for large language models.
\newblock In \emph{2024 4th International Conference on Artificial Intelligence, Robotics, and Communication (ICAIRC)}, 224--231. IEEE.

\bibitem[{Liao et~al.(2024)Liao, Herold, Khadivi, and Monz}]{liao2024apiq}
Liao, B.; Herold, C.; Khadivi, S.; and Monz, C. 2024.
\newblock {ApiQ}: Finetuning of 2-bit quantized large language model.
\newblock In \emph{Proceedings of the 2024 Conference on Empirical Methods in Natural Language Processing}, 20996--21020.

\bibitem[{Lu(2001)}]{dsp_fir}
Lu, W.-S. 2001.
\newblock Design of FIR filters with discrete coefficients: a semidefinite programming relaxation approach.
\newblock In \emph{ISCAS 2001. The 2001 IEEE International Symposium on Circuits and Systems}, volume~2, 297--300 vol. 2.

\bibitem[{Matej et~al.(1999)Matej, Vardi, Herman, and Vardi}]{Matej1999}
Matej, S.; Vardi, A.; Herman, G.~T.; and Vardi, E. 1999.
\newblock \emph{Binary Tomography Using Gibbs Priors}, 191--212.
\newblock Boston, MA: Birkh{\"a}user Boston.
\newblock ISBN 978-1-4612-1568-4.

\bibitem[{McClellan and Parks(1973)}]{remez}
McClellan, J.; and Parks, T. 1973.
\newblock A unified approach to the design of optimum FIR linear-phase digital filters.
\newblock \emph{IEEE Transactions on Circuit Theory}, 20(6): 697--701.

\bibitem[{Morozov, Zheltkov, and Osinsky(2024)}]{AAM}
Morozov, S.; Zheltkov, D.; and Osinsky, A. 2024.
\newblock Accelerated alternating minimization algorithm for low-rank approximations in the Chebyshev norm.
\newblock arXiv:2410.05247.

\bibitem[{{NVIDIA Corporation}(2025)}]{nccl}
{NVIDIA Corporation}. 2025.
\newblock NVIDIA Collective Communications Library (NCCL) Documentation.
\newblock Accessed: 2025-07-31.

\bibitem[{Oppenheim and Schafer(2009)}]{oppenheim_schafer}
Oppenheim, A.~V.; and Schafer, R.~W. 2009.
\newblock \emph{Discrete–Time Signal Processing}.
\newblock Prentice Hall, 3rd edition.

\bibitem[{Pan, Sidky, and Vannier(2009)}]{fbp}
Pan, X.; Sidky, E.~Y.; and Vannier, M. 2009.
\newblock Why do commercial {CT} scanners still employ traditional, filtered back-projection for image reconstruction?
\newblock \emph{Inverse Problems}, 25(12): 123009.

\bibitem[{Pedano et~al.(2016)Pedano, Flanders, Scarpace, Mikkelsen, Eschbacher, Hermes, Sisneros, Barnholtz-Sloan, and Ostrom}]{brainct}
Pedano, N.; Flanders, A.~E.; Scarpace, L.; Mikkelsen, T.; Eschbacher, J.~M.; Hermes, B.; Sisneros, V.; Barnholtz-Sloan, J.; and Ostrom, Q. 2016.
\newblock The Cancer Genome Atlas Low Grade Glioma Collection (TCGA-LGG).

\bibitem[{Poursoltani and Delage(2022)}]{Poursoltani}
Poursoltani, M.; and Delage, E. 2022.
\newblock Adjustable Robust Optimization Reformulations of Two-Stage Worst-Case Regret Minimization Problems.
\newblock \emph{Oper. Res.}, 70(5): 2906–2930.

\bibitem[{Qiu et~al.(2017)Qiu, Xu, Xu, and Tan}]{Continuous_minimax2}
Qiu, X.; Xu, J.-X.; Xu, Y.; and Tan, K.~C. 2017.
\newblock A new differential evolution algorithm for minimax optimization in robust design.
\newblock \emph{IEEE Transactions on Cybernetics}, 48(5): 1355--1368.

\bibitem[{Rabiner(1971)}]{Rabiner}
Rabiner, L. 1971.
\newblock Techniques for Designing Finite-Duration Impulse-Response Digital Filters.
\newblock \emph{IEEE Transactions on Communication Technology}, 19(2): 188--195.

\bibitem[{Ropke and Pisinger(2006)}]{ALNS}
Ropke, S.; and Pisinger, D. 2006.
\newblock An Adaptive Large Neighborhood Search Heuristic for the Pickup and Delivery Problem with Time Windows.
\newblock \emph{Transportation Science}, 40(4): 455--472.

\bibitem[{Sainz et~al.(2008)Sainz, Herrero, Armengol, and Veh{\'\i}}]{Continuous_minimax1}
Sainz, M.~{\'A}.; Herrero, P.; Armengol, J.; and Veh{\'\i}, J. 2008.
\newblock Continuous minimax optimization using modal intervals.
\newblock \emph{Journal of Mathematical Analysis and Applications}, 339(1): 18--30.

\bibitem[{Shang et~al.(2023)Shang, Yuan, Wu, and Dong}]{shang2023pb}
Shang, Y.; Yuan, Z.; Wu, Q.; and Dong, Z. 2023.
\newblock {PB-LLM}: Partially binarized large language models.
\newblock In \emph{International Conference on Learning Representations}.

\bibitem[{Shao et~al.(2024)Shao, Chen, Zhang, Xu, Zhao, Li, Zhang, Gao, Qiao, and Luo}]{shao2023omniquant}
Shao, W.; Chen, M.; Zhang, Z.; Xu, P.; Zhao, L.; Li, Z.; Zhang, K.; Gao, P.; Qiao, Y.; and Luo, P. 2024.
\newblock {OmniQuant}: Omnidirectionally calibrated quantization for large language models.
\newblock In \emph{The Twelfth International Conference on Learning Representations}.

\bibitem[{Ta and Le-Nhat(2008)}]{Hung}
Ta, H.~Q.; and Le-Nhat, T. 2008.
\newblock Design of FIR filter with discrete coefficients based on Mixed Integer Linear Programming.
\newblock In \emph{2008 9th International Conference on Signal Processing}, 9--12.

\bibitem[{van Aarle et~al.(2016)van Aarle, Palenstijn, Cant, Janssens, Bleichrodt, Dabravolski, Beenhouwer, Batenburg, and Sijbers}]{astra2}
van Aarle, W.; Palenstijn, W.~J.; Cant, J.; Janssens, E.; Bleichrodt, F.; Dabravolski, A.; Beenhouwer, J.~D.; Batenburg, K.~J.; and Sijbers, J. 2016.
\newblock Fast and flexible X-ray tomography using the ASTRA toolbox.
\newblock \emph{Opt. Express}, 24(22): 25129--25147.

\bibitem[{{van Aarle} et~al.(2015){van Aarle}, Palenstijn, {De Beenhouwer}, Altantzis, Bals, Batenburg, and Sijbers}]{astra1}
{van Aarle}, W.; Palenstijn, W.~J.; {De Beenhouwer}, J.; Altantzis, T.; Bals, S.; Batenburg, K.~J.; and Sijbers, J. 2015.
\newblock The ASTRA Toolbox: A platform for advanced algorithm development in electron tomography.
\newblock \emph{Ultramicroscopy}, 157: 35--47.

\bibitem[{Wang et~al.(2024)Wang, Liu, Feng, Ding, and Ding}]{wang2024fp4}
Wang, J.; Liu, H.; Feng, D.; Ding, J.; and Ding, B. 2024.
\newblock {FP4}-quantization: Lossless 4bit quantization for large language models.
\newblock In \emph{2024 IEEE International Conference on Joint Cloud Computing (JCC)}, 61--67. IEEE.

\bibitem[{Wang et~al.(2004)Wang, Bovik, Sheikh, and Simoncelli}]{SSIM}
Wang, Z.; Bovik, A.; Sheikh, H.; and Simoncelli, E. 2004.
\newblock Image quality assessment: from error visibility to structural similarity.
\newblock \emph{IEEE Transactions on Image Processing}, 13(4): 600--612.

\bibitem[{Wesson, Ochshorn, and Land(2009)}]{bdstop}
Wesson, K.; Ochshorn, R.; and Land, B. 2009.
\newblock Low-cost, high-fidelity, adaptive cancellation of periodic 60 Hz noise.
\newblock \emph{Journal of Neuroscience Methods}, 185: 50--5.

\bibitem[{Xiao et~al.(2023)Xiao, Lin, Seznec, Wu, Demouth, and Han}]{xiao2023smoothquant}
Xiao, G.; Lin, J.; Seznec, M.; Wu, H.; Demouth, J.; and Han, S. 2023.
\newblock {SmoothQuant}: Accurate and efficient post-training quantization for large language models.
\newblock In \emph{International Conference on Machine Learning}, 38087--38099. PMLR.

\bibitem[{Yi and Neykov(2024)}]{yi_Linf}
Yi, Y.; and Neykov, M. 2024.
\newblock Non-asymptotic bounds for the $\ell_\infty$ estimator in linear regression with uniform noise.
\newblock \emph{Bernoulli}, 30.

\bibitem[{Zhang et~al.(2022)Zhang, Roller, Goyal, Artetxe, Chen, Chen, Dewan, Diab, Li, Lin, Mihaylov, Ott, Shleifer, Shuster, Simig, Koura, Sridhar, Wang, and Zettlemoyer}]{optmodel}
Zhang, S.; Roller, S.; Goyal, N.; Artetxe, M.; Chen, M.; Chen, S.; Dewan, C.; Diab, M.; Li, X.; Lin, X.~V.; Mihaylov, T.; Ott, M.; Shleifer, S.; Shuster, K.; Simig, D.; Koura, P.~S.; Sridhar, A.; Wang, T.; and Zettlemoyer, L. 2022.
\newblock OPT: Open Pre-trained Transformer Language Models.
\newblock arXiv:2205.01068.

\bibitem[{Zhu et~al.(2024)Zhu, Li, Liu, Ma, and Wang}]{zhu2024survey}
Zhu, X.; Li, J.; Liu, Y.; Ma, C.; and Wang, W. 2024.
\newblock A survey on model compression for large language models.
\newblock \emph{Transactions of the Association for Computational Linguistics}, 12: 1556--1577.

\end{thebibliography}
